\documentclass[review]{elsarticle}
\usepackage{lineno}

\graphicspath{ {./figures/} }
\usepackage{float}
\usepackage{verbatim} 
\usepackage{adjustbox}
\usepackage{array,multirow}
\usepackage{rotating}
\usepackage{graphicx}
\usepackage{booktabs}
\usepackage{makecell}
\usepackage[inline]{enumitem}
\usepackage{amsmath}

\newcommand{\eg}{e.g., }
\newcommand{\ie}{i.e., }

\newcommand{\figref}[1]{Fig.~\ref{#1}}    
\newcommand{\Figref}[1]{Figure~\ref{#1}}  
\newcommand{\tabref}[1]{Table~\ref{#1}}

\newcommand{\secref}[1]{Section~\ref{#1}}

\newcommand{\Fone}{{F$_{1}$}}
\newcommand{\Fonec}{{F$_{1c}$}}
\newcommand{\Fones}{{F$_{1s}$}}

\newcommand{\clstag}{{\tt{[CLS]}}}
\newcommand{\septag}{{\tt{[SEP]}}}

\usepackage[colorlinks]{hyperref}
\hypersetup{colorlinks = true,linkcolor = blue,anchorcolor =red,citecolor = blue,filecolor = red,urlcolor = red,pdfauthor=author}

\newcommand{\giannis}[1]{\textcolor{black}{#1}}
\newcommand{\giannisrtwo}[1]{\textcolor{black}{#1}}
\newcommand{\xiangyu}[1]{\textcolor{black}{#1}}
\newcommand{\nikos}[1]{\textcolor{black}{#1}}

\newcommand{\yang}[1]{\textcolor{black}{#1}}

\newcommand{\cmark}{\ding{51}}%
\newcommand{\xmark}{\ding{55}}%

\restylefloat{figure}
\restylefloat{table}

\biboptions{authoryear}

\usepackage{tikz,xcolor,hyperref}

\definecolor{lime}{HTML}{A6CE39}
\DeclareRobustCommand{\orcidicon}{
	\begin{tikzpicture}
	\draw[lime, fill=lime] (0,0) 
	circle [radius=0.16] 
	node[white] {{\fontfamily{qag}\selectfont \tiny ID}};
	\draw[white, fill=white] (-0.0625,0.095) 
	circle [radius=0.007];
	\end{tikzpicture}
	\hspace{-2mm}
}

\foreach \x in {A, ..., Z}{\expandafter\xdef\csname orcid\x\endcsname{\noexpand\href{https://orcid.org/\csname orcidauthor\x\endcsname}
			{\noexpand\orcidicon}}
}

\makeatletter
\def\ps@pprintTitle{%
   \let\@oddhead\@empty
   \let\@evenhead\@empty
   \let\@oddfoot\@empty
   \let\@evenfoot\@oddfoot
}
\makeatother

\begin{document}
\begin{frontmatter}


\author[vub,imec]{Xiangyu Yang\corref{cor1}}
\ead{xyanga@etrovub.be}
\author[vub,imec]{Giannis Bekoulis}
\ead{gbekouli@etrovub.be}
\author[vub,imec]{Nikos Deligiannis\orcidA{}}
\ead{ndeligia@etrovub.be}
\cortext[cor1]{Corresponding author}
\address[vub]{ETRO, Vrije Universiteit Brussel, Pleinlaan 2, B-1050 Brussels, Belgium }
\address[imec]{imec, Kapeldreef 75, B-3001 Leuven, Belgium}

\title{Traffic Event Detection as a Slot Filling Problem}

\begin{abstract}
Twitter is a social media platform with increasing popularity, \nikos{where} users can share their thoughts around different topics through short messages. Studies have shown that Twitter streams contain rich information regarding traffic-related topics. 
\giannis{By using this large amount of information, we can build smart traffic event detection systems to help people avoid traffic jams, accidents, etc.}
In this paper, we introduce the new problem of extracting fine-grained traffic information from Twitter streams by also making publicly available the two (constructed) traffic-related datasets from Belgium and the Brussels capital region.  In particular, we \yang{experiment with} several models to identify (i) whether a tweet is traffic-related or not, and (ii) in the case that the tweet is \giannis{traffic-related} to identify more fine-grained information regarding the event (e.g., the type of the event, where the event happened). To do so, we frame (i) the problem of identifying whether a tweet is a traffic-related event or not as a text classification \giannisrtwo{sub}task, and (ii) the problem of identifying more fine-grained traffic-related information as a slot filling \giannisrtwo{sub}task, where fine-grained information (e.g., where an event has happened) is represented as a slot/entity of a particular type.
\giannisrtwo{We propose the use of several methods that process the two subtasks either separately or in a joint setting, and we evaluate the effectiveness of the proposed methods for solving the traffic event detection problem. 
Experimental results indicate that the proposed architectures achieve high performance scores (\ie more than 95\% in terms of \Fone~score) on the constructed datasets for both of the subtasks (\ie text classification and slot filling) even in a transfer learning scenario. In addition, by incorporating tweet-level information in each of the tokens comprising the tweet (for the BERT-based model) can lead to a performance improvement for the joint setting.}
Our datasets and code will be available on GitHub upon acceptance of the manuscript.

\end{abstract}

\begin{keyword}
traffic event detection \sep slot filling \sep text classification \sep deep learning
\end{keyword}

\end{frontmatter}

\section{Introduction}
\label{introduction}

\noindent 
\giannisrtwo{In smart cities, digital technologies are exploited to improve the citizens' quality of life. For instance, this can be achieved by identifying efficient ways to heat buildings, upgrading the waste disposal facilities, and also building smart city transport networks.
In several cities around the world, people experience traffic conditions, such as traffic jams or even accidents that can negatively impact people’s life~\citep{congest, traffic_issue}.}
By building an intelligent system that can provide useful online traffic information, traffic problems can potentially be mitigated \yang{and it can help facilitate the building of a smart city.} For example, based on real-time traffic information,  travelers can find the best route to their destination\giannis{,} which is less expensive and less time consuming; commuters can avoid traffic jams; the traffic management centers can easily monitor traffic flows and inform police once there is an incident, so that traffic can be quickly restored. Thus, it is critical to build a smart system that can detect traffic events in 
\giannisrtwo{every city}.
This research focuses on building a novel traffic event detection system using social media that focuses, \yang{especially}, in the Brussels capital region, but we showcase that it can perform well also in the entire country (i.e., Belgium). 
\giannisrtwo{Although the traffic detection system presented in our research focuses on the Brussels capital region (and in Belgium) and is in Dutch, the principles of the system can be easily extended to other cities.}

Social media allow users to easily access and provide information regarding anything at any time. Twitter, one of the most popular social media platform\giannis{s}, has around 300 million monthly active users. The reason that Twitter is so popular is that its users are allowed to share their thoughts through short texts (which are called tweets), which is a straightforward way for the users to connect the one with the other. With such a large number of users, Twitter generates a large amount of data; in particular, 500 million tweets every day~\citep{tweets}. Since a large amount of content is produced, a lot of recent research has \giannis{focused on} identifying important patterns from people's daily life. This research includes  extracting events from social media posts (e.g.,  \giannis{disasters, emergencies~\citep{castillo2016big}}), extract sub-events from other events (e.g., a football match~\citep{bekoulis-etal-2019-sub}). 

\nikos{T}weets also contain a lot of information regarding the traffic conditions in a particular region. For instance, there are users or even \yang{accounts of} official channels that are tweeting during the day about possible traffic jams, accidents, and so on. Thus, it is important \giannis{for people} to exploit the traffic-related information that \giannis{comes} from Twitter streams in order to build smart traffic event detection systems \giannis{and avoid the aforementioned issues (\eg traffic jams)}. According to the work of~\citet{dabiri2019developing}, traffic events can be divided into two categories: \giannis{\textit{recurring}} and \giannis{\textit{non-recurring}} events. Recurring events are those that can be easily predicted by looking into historical data\giannis{,} such as daily rush hours. Non-recurring events refer to unpredictable events such as traffic accidents (car crashes), weather caused issues, and natural disasters. The traditional way to detect those \yang{\textit{non-recurring}} traffic events 
is to use sensors and cameras installed in the city. However, \nikos{deploying such devices everywhere in the city is costly and technically challenging}. 
\giannisrtwo{Exploiting the information coming from the Twitter stream can provide complementary information for building a smart traffic event detection system.}

In this work, we focus on detecting traffic-related events in Belgium and the Brussels capital region from the Twitter stream.
To do so, we collect and annotate two traffic-related Dutch Twitter datasets from Belgium and the Brussels capital region. Our datasets contain tweet-level information about whether a tweet is traffic-related or not\nikos{. In case a tweet is traffic-related, more fine-grained information is included, that is, 
the time (``when''),  the location (``where”), the type (``what”), and the ``consequence” of the reported traffic event (see the definition of the slot filling subtask in~\secref{3.1} for more details).}
We can approach this problem as a series of two subtasks, namely 
\begin{enumerate*}[label=(\roman*)]
\item text classification and\label{step1}
\item slot filling.~\label{step2}
\end{enumerate*}
\giannis{The two subtasks can be considered either as two independent subtasks} \nikos{or can be addressed jointly.}
The goal of \giannis{subtask}~\ref{step1} is to assign a set of predefined categories \giannis{(\ie traffic-related and non-traffic-related) to a textual document (\ie a tweet in our case)}. 
For \giannis{sub}task~\ref{step2}, the goal is to identify \giannis{text spans inside the traffic-related tweets that answer the predefined questions} defined above (\ie ``when", ``where", ``what", and ``consequence"). 
Researchers have identified the benefit of training related subtasks together in a joint setting since the interactions between the subtasks are taken into account (see for instance prior work on multitask learning~\citep{caruana1997multitask}, entity recognition and relation extraction~\citep{miwa2016end,bekoulis2018joint}, \giannis{and} tree-structured prediction~\citep{bekoulis2018attentive}).

To summarize, the key contributions of our work are as follows:
\begin{itemize}
    
    \item We define \nikos{a} new traffic event detection problem and publish two Dutch datasets \giannis{(one with tweets from the Brussels capital region and one with tweets from Belgium) annotated with class- and span-level information for each tweet (as described above). That way, \nikos{we promote} the research of traffic event detection from Twitter streams.}
    \item \giannis{We propose to solve the problem of detecting traffic-related events from Twitter streams by a series of two subtasks, namely, text classification and slot filling. 
    We experiment with several architectures (\eg BERT-based models) and we solve each subtask either independently or in a joint setting.}
    \giannisrtwo{Furthermore, we modify the joint BERT-based model by incorporating the entire information of the tweet into each of its composing tokens. This model is able to outperform all other models in the joint setting.}
    \item We carry out extensive experiments, and \giannis{our experimental study indicates} the effectiveness of our BERT-based methods \giannis{over other studied baseline models (\eg LSTM-based models)} for detecting traffic-related events on Twitter.
    
\end{itemize} 

The rest of this work is organized as follows.~\secref{relatedwork} reviews the related work on traffic event detection systems \giannis{from Twitter streams}.~\secref{pf} introduces the newly defined task of identifying fine-grained traffic information from traffic-related tweets and describes the annotation process.~\secref{method} describes the \giannis{various \yang{proposed} architectures for solving the task defined in~\secref{pf}}.~\secref{experiments} describes the experimental setup for the proposed methods, introduces the experimental evaluation for the tasks, and showcases the performance of the proposed models.~\secref{conclusion} concludes our work and discusses future \giannis{research directions}. 

\section{Related Work}
\label{relatedwork}

\noindent A lot of research in the NLP community has been focused \nikos{on} detecting events from social media (\eg \cite{Naseem2019DeepCE, bekoulis-etal-2019-sub, dabiri2019developing, muller2020covid}). There are two types of event detection, \nikos{namely, specified and unspecified,} as indicated in the work of~\citet{saeed2019s}. \nikos{In} specified event detection, the event types are \nikos{determined upfront} and there is a wide range of event types, such as earthquake~\citep{sakaki2010earthquake, sakaki2012tweet}, traffic~\citep{ali2021traffic, dabiri2019developing}, epidemic~\citep{zong2020extracting, muller2020covid}, sports news~\citep{adedoyin2016rule}, etc. \nikos{In} unspecified event detection there is no prior information about the event types. The techniques used for event detection can be classified into two categories: unsupervised and supervised approaches. Unsupervised approaches usually involve various clustering algorithms (\eg cluster tweets that contain \giannis{the} top-$k$ bursting keywords from Twitter streams~\citep{li2012twevent}). Supervised approaches are mainly used for specified event type detection and the task is mostly framed as a text classification~\citep{dabiri2019developing, alomari2019road} or as a slot filling problem~\citep{zong2020extracting}. In this paper, we propose new models for solving the newly defined task of identifying fine-grained information from traffic-related tweets (as detailed briefly in~\secref{introduction} and described in more detail in~\secref{pf}). We formulate the task of identifying whether a tweet is traffic-related or not into a text classification problem, and the task of identifying fine-grained information as a slot filling problem. In literature, the text classification and slot filling tasks are often handled in a joint setting and the joint task is named \giannis{as} joint intent detection and slot filling~\citep{weld2021survey}. The intent is the intention of the speaker in an utterance \giannis{and the prediction of the intent is mostly treated as a text classification task~\citep{Larson2019AnED}}. 
\giannis{In the following subsections, we} present related work on the task of traffic-related event detection \giannis{(using text classification methods}, see~\secref{ted}), recent work on \giannis{the slot filling task} (see~\secref{sf}), and relevant work on the task of joint intent detection and slot filling \giannis{(see~\secref{joint_idsf})}.

\subsection{Traffic Event Detection}
\label{ted}

\noindent The traffic event detection problem is mainly approached as a text classification \giannis{task} in the literature. Machine learning based methods  
\giannis{have been exploited}
to tackle the problem, 
\giannis{both} the traditional (\eg Support Vector Machine (SVM)~\citep{DAndrea2015RealTimeDO, Salas2017IncidentDU}, Naïve Bayes (NB)~\citep{Gu2016FromTT}, Decision Tree (DT)~\citep{Wongcharoen2016TwitterAO}) and the deep learning ones (\eg Convolutional Neural Networks (CNNs)~\citep{dabiri2019developing, Chen2019DetectingTI}, Recurrent Neural Networks (RNNs)~\citep{ali2021traffic, dabiri2019developing, Chen2019DetectingTI}).

\noindent \textbf{Traditional Machine Learning}:~\citet{DAndrea2015RealTimeDO} proposed a real-time traffic event monitoring system that can fetch tweets from Italian Twitter streams and classify them into the appropriate classes: traffic and non-traffic. They used the Inverse Document Frequency (IDF) technique to represent tweets as features and built \giannis{an} SVM classifier for classification. Instead of using the IDF technique,~\citet{Salas2017IncidentDU} explored the use of n-grams as features \giannis{to} an SVM classifier to detect traffic incidents.~\citet{Gu2016FromTT} presented a real-time architecture to detect traffic incidents \giannis{in Twitter streams}. In particular, they established a dictionary 
\giannis{of keywords and their combinations to}
infer traffic incidents. Based on \giannis{that} dictionary, a tweet is represented as a high dimensional binary vector. Those binary vectors are then fed into a NB classifier to identity whether the corresponding tweets are traffic incidents or not.~\citet{Wongcharoen2016TwitterAO} proposed a model to detect \giannis{the} congestion severity level\giannis{s from Twitter streams}. \giannis{In that work,} they considered four attributes (\giannis{\ie} day of \giannis{the} week, hours of the day, minutes of the hour, and tweets density) to construct a C4.5 DT for congestion severity level prediction.~\citet{Wang2017RealTimeTE} introduced a Latent Dirichlet Allocation (LDA) model called tweet-LDA to extract traffic alert and warning topics from Twitter in real-time.

\noindent \textbf{Deep Learning}:~\citet{zhang2018deep} developed a traffic accident detection system that uses tokens that are relevant to traffic (\eg accident, car, and crash) as features to train a Deep Belief Network (DBN). 
\giannis{The selection of tokens is based on a coefficient that} measures the association between the labels (accident or not) and the tokens.
~\citet{Chen2019DetectingTI} built a binary classification system to detect traffic-related information from Weibo \giannis{(a Chinese social media platform)}. 
They applied continu\giannis{ous}  bag-of-words (CBOW) to learn word embeddings to represent words in microblogs. 
Then, \nikos{they used} the learned word embeddings as input to CNNs, Long Short-Term Memory (LSTM) \giannis{networks}, and \giannis{their combined LSTM-CNN architecture} to detect traffic-related microblogs.
~\citet{dabiri2019developing} proposed to address the traffic event detection problem on Twitter as a text classification problem using deep learning architectures. In particular, in \giannis{their} work, they first collected and labeled a traffic-related Twitter dataset from the USA, which contains three classes: non-traffic (events that are not related to traffic), traffic incident (non-recurring events such as car crashes, traffic signal problem, and disabled vehicles), and traffic information \& condition (recurring events such as traffic congestion, daily rush hours, and traffic delays). After that, they used pre-trained word embedding models (word2vec~\citep{Mikolov2013EfficientEO} and FastText~\citep{Bojanowski2017EnrichingWV}) to get tweet representations. CNNs and RNNs were deployed on top of the \giannis{word embeddings layer} 
to extract traffic-related tweets.~\citet{ali2021traffic} presented an architecture to detect traffic accident\giannis{s} and analyze the \giannis{traffic} condition\giannis{s directly} from social networking data. They first collected traffic information from \giannis{the} Twitter and Facebook \giannis{networking platforms} by \giannis{using} a query-based search engine. Then, \giannis{Ontologies} 
and Latent Dirichlet Allocation (OLDA) 
\nikos{were used} to automatically label each sentence \giannis{with either the traffic or the non-traffic class labels.} 
Finally, they used FastText with bidirectional LSTM\giannis{s} (BiLSTM\giannis{s}) to detect domain-specific event types (\eg traffic accidents and traffic jam) and predict user sentiments (\ie positive, neutral, or negative) towards those traffic events.

\yang{Previous work on traffic event detection using social media mainly focuses on classifying tweets into two classes, traffic-related and non-related~\citep{DAndrea2015RealTimeDO,Salas2017IncidentDU,Gu2016FromTT,zhang2018deep,Chen2019DetectingTI,dabiri2019developing}. 
\giannis{Although identifying whether a tweet is traffic-related or not is important, it is also crucial to know more precise information regarding a particular event (as reported in the Twitter stream).}
\giannis{For instance, we are interested to know ``where'' or ``when'' a reported traffic event happened.}
\giannis{I}n this paper, we propose to process the traffic event detection problem as \giannisrtwo{a series of} two subtasks: (i) determining whether a tweet is traffic-related or not (which we treat as a text classification problem), and (ii) detecting fine-grained information \giannis{(\eg where)} from tweets \giannis{(which we treat as a slot filling problem)}. 
\giannis{The fine-grained information (\eg ``where'' or ``when'' an event has happened) could help us decide the nature of the event (\eg whether it is traffic-related or not). Moreover, in the case that the event is traffic-related it could also help us to decide about whether we should identify text spans for the slot filling task. We conduct extensive experiments and we study the two subtasks either separately or in a joint setting to identify whether there is a benefit by explicitly sharing the layers of the neural network between the subtasks.}
The way \giannis{that} we formulate the traffic event detection problem \giannis{has not been studied before} in the traffic event detection domain, and we hope \giannisrtwo{that this could boost future research on traffic event detection using social media.}}

\subsection{Slot Filling}
\label{sf}
\noindent In Natural Language Understanding (NLU), slot filling 
\nikos{is a task whose goal is to identify spans of text (\ie the start and the end position)} that belong to predefined classes \giannis{directly} from raw text.
A span is a semantic unit that consists of a set of contiguous words\giannis{. For example, the span} ``The Lord of the Rings" refers to the famous novel\giannis{/movie}. 
\nikos{The slot filling task is mainly used in the context of dialog systems where the aim is to retrieve the required information (\ie slots) out of the textual description of the dialog}. 
For instance, when a user queries \giannis{about} top-rated novels, the dialog system should be able to retrieve relevant novel titles and \giannis{the corresponding} ratings. 

\nikos{Slot filling is usually formulated as a sequence labeling task and neural network based models have mainly been proposed for solving it}. In particular,~\citet{vu} proposed a bidirectional sequential CNN model that predicts the label for each slot by 
\giannis{taking into account the context (\ie previous and future) words with respect to the current word and the current word itself.}~\citet{kurata} developed the encoder-labeler LSTM that first uses the encoder LSTM to encode the entire input sequence into a fixed length vector\giannis{.}
\giannis{This vector} is then passed as the initial state to the labeler LSTM \giannis{to perform the sequence labeling task}. 
This model is able to predict slot labels while taking into account the \giannis{whole information of the input sequence}.~\citet{Zhu2017EncoderdecoderWF} introduced the BiLSTM-LSTM, an encoder-decoder model that encodes the input sequence using a BiLSTM and decodes the encoded information \giannis{using} a \giannis{unidirectional} LSTM. 
\giannis{They also designed a so-called focus mechanism that is able to address the alignment limitation of attention mechanisms (\ie cannot operate with a limited amount of data) for sequence labeling.}
~\cite{Zhao2018ImprovingSF} presented a sequence-to-sequence (Seq2Seq) model along with a pointer network to improve the slot filling performance. To predict slot values, the model learns to \giannis{either} copy a word (which may be out-of-vocabulary (OOV)) through a pointer network, or generate a word within the vocabulary through an attentional Seq2Seq model.~\citet{Korpusik2019ACO} compared a set of neural networks (CNN, RNN, BiLSTM, and Bidirectional Encoder Representations from Transformers (BERT)) on slot filling tasks. Their results indicate that the BERT\giannis{-based models} outperform the other \giannis{studied architectures}.~\giannis{Recently, }\citet{zong2020extracting} published the COVID-19 Twitter Event Corpus, which has 7,500 annotated tweets \giannis{and includes} five event types (TESTED POSITIVE, TESTED NEGATIVE, CAN NOT TEST, DEATH, and CURE\&PREVENTION). For each event type, a set of slot types \giannis{is} predefined for slot filling tasks (\eg for the TESTED POSITIVE event, the goal is to identify slot types like ``who" (\ie who was tested positive), ``age" (\ie the age of the person tested positive), and ``gender" (\ie the gender of the person tested positive)). They proposed a BERT-based model that treats each slot filling task in each event type as a binary classification problem. Compared to their model,
\giannisrtwo{in~\citet{Yang2020imecETROVUBAW}},
we proposed a multilabel BERT-based model that jointly trains all the slot types for a single event and achieves \giannis{improved} slot filling performance.
\giannisrtwo{In this paper, we modify existing slot filling techniques, and we apply them in the context of traffic event detection from Twitter streams.}
 
\subsection{Joint Intent Detection and Slot Filling}
\label{joint_idsf}

\noindent \giannis{The tasks of intent detection and slot filling have also been studied in a joint setting. }
Given an utterance, intent detection aims to identify \giannis{the intention of the user (\eg book a restaurant)} \giannis{and the} slot filling \giannis{task} focuses on extracting text spans that are relevant to \giannis{that} intention \giannis{(\eg place of the restaurant, timeslot)}. 
\giannis{By training the two tasks simultaneously (\ie in a joint setting), the model is able to learn the inherent relationships between the two tasks of intention detection and slot filling.} \nikos{This approach can further improve the overall performance of the joint task and the performance of each independent subtask.} \giannis{The benefit of training tasks simultaneously is also indicated in~\secref{introduction} (interactions between subtasks are taken into account) and more details on the benefit of multitask learning can also be found in the work of~\cite{caruana1997multitask}.}   
\giannis{A} detailed survey on \giannis{studying the two tasks of intent detection and slot filling in a joint setting} can be found in the work of~\citet{weld2021survey}. 
\giannis{The two types of methods that are mainly exploited when solving the two tasks simultaneously are the RNN-based and the attention-based approaches.}

\noindent \textbf{RNN-based\giannis{:}}~\citet{Zhou2016AHL} proposed a hierarchical LSTM model which has two LSTM layers. The final hidden state of the bottom LSTM layer is used for intent detection, while that of the top LSTM layer with a softmax classifier is used to label 
\giannis{the tokens of}
the input sequence.~\citet{HakkaniTr2016MultiDomainJS} developed a single BiLSTM model that concatenates the hidden states of the forward and \giannis{the} backward layers of an input token and passes \giannis{these} concatenated features to 
\giannis{a softmax classifier}
to predict the slot label for that token. A special tag is added \giannis{at the end of} the input sequence for capturing the 
\giannis{context of the whole sequence and detecting the class of the intent.}
\citet{Zhang2016AJM} proposed a bidirectional gated recurrent unit (GRU) architecture \giannis{that operates in a similar way to the work of~\citet{HakkaniTr2016MultiDomainJS} for labeling the slots. However, the intent detection is done in a different way since \citet{Zhang2016AJM} use a max-pooling layer for all the hidden states, and then they apply a softmax function on top of the max-pooling layer.}
~\citet{Firdaus2018ADL} introduced an ensemble model that feeds the outputs of a BiLSTM and a BiGRU separately into two multi-layer perceptrons (MLP)
. The outputs of the MLPs are \giannis{concatenated} and a softmax classifier is used for predicting the intent and \giannis{the} slots simultaneously. 

\noindent \textbf{Attention-based:} 
\giannis{Attention mechanisms have also been exploited for jointly learning the relationships between the two studied subtasks.}
\giannis{In particular,}~\citet{Liu2016AttentionBasedRN} proposed an attention-based bidirectional RNN (BRNN) model that takes the weighted sum of the concatenation of the forward and \giannis{the} backward hidden states as an input to predict the intent and the slots.~\citet{Li2018ASM} proposed \giannis{the use of} a BiLSTM model with the self-attention mechanism~\citep{vaswani2017attention} and \giannis{a} gate mechanism to \giannis{solve} the joint task. One self-attention mechanism is used \giannis{at the} words and \giannis{the} characters \giannis{level of} the input sequence to obtain \giannis {a} semantic representation of \giannis{the input}. Then, the\giannis{se} representations are fed into a BiLSTM, and the final hidden state is \giannis{then} used for intent detection. Another self-attention layer is applied between the intermediate states of the BiLSTM, and the intermediate states are combined with \giannis{the predicted intent for labeling the slots}.
\citet{goo2018slot} introduced an attention-based slot-gated BiLSTM model. 
\giannis{In that model, the embeddings of the input sentence are fed into a BiLSTM, and then a weighted sum of the BiLSTM intermediate states (\ie the slot context vector) is used for predicting the slots.} 
The final state of the BiLSTM (\ie the intent context vector) \giannis{is used for predicting the intent}. A slot gate is added to combine the slot context vector with the intent context vector, and the combined vector is then feed into a softmax to predict the current slot label.~\citet{chen} proposed a joint BERT model for \giannis{solving the} joint task \giannis{of intent detection and slot filling}. The model predicts the intent based on the final hidden state of the~\clstag~token, and the final hidden state of each token is used \giannis{for predicting} the slot label\giannis{s}. The two tasks are trained jointly by using a joint loss \giannis{(\ie one for each subtask)}. Our joint model is conceptually related to \giannis{that of}~\citet{chen}.
\giannis{Unlike the work of~\cite{chen} that does not exploit the use of the intent information, we incorporate intent information into each token for improving the performance of the slot filling subtask. This also improves the overall performance of the joint task (\ie intent detection and slot filling). }

\section{Problem Formulation and Datasets}
\label{pf}

\noindent 
\nikos{In this section, we define the traffic event detection problem from Twitter streams and explain that this problem can be addressed by the two subtasks of text classification and slot filling.} 
\giannis{We then present the way that the two datasets (\ie the one for Belgium and the other for the Brussels capital region) have been constructed (\ie data collection, annotation process).}


\subsection{Subtasks}
\label{3.1}
\begin{figure}
    \centering
    \resizebox{0.9\columnwidth}{!}{%
    \includegraphics[scale=0.55]{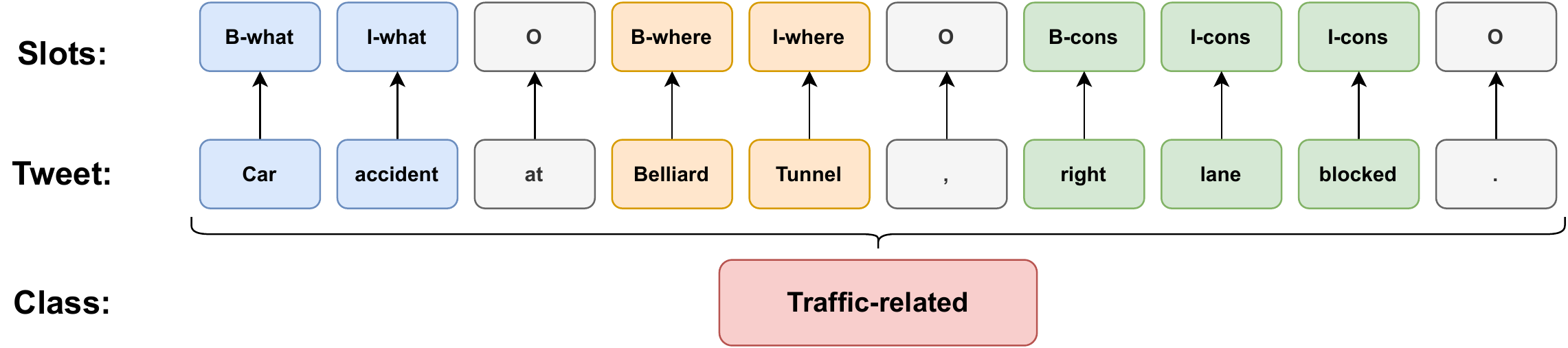}
    \caption{
    \giannis{An example tweet, where the two subtasks of text classification and slot filling are illustrated. The output for the text classification problem is the class label of the tweet (\ie traffic-related). The output for the slot filling problem is for each token the slot filling label (\eg ``what'', ``where'') encoded using the BIO encoding scheme. }
    }
    \label{fig:subtasks}}
\end{figure}

\noindent \nikos{By formulating the problem into a series of the two subtasks (text classification and slot filling), we give an answer to two questions: (i) whether the given tweet is traffic-related (or not), and (ii) whether more precise/fine-grained information can be identified regarding a traffic-related event (from the corresponding tweet).} The two subtasks for \giannis{a particular} tweet are illustrated in~\figref{fig:subtasks}. 


\noindent \textbf{Text Classification\giannis{:}} The goal of this subtask is to distinguish traffic-related tweets from non-traffic-related tweets. Since there are two classes, this subtask, in its essence, is a binary classification task
\giannis{. That is,} given a tweet \textit{t}, the classification model $f(t)\rightarrow \left \{ 0,1 \right \}$ predicts whether a tweet is traffic-related or not, where 1 means \textit{traffic-related}, \giannis{and} 0 means \textit{non-traffic-related}.

\noindent \textbf{Slot Filling\giannis{:}} This subtask aims at extracting fine-grained events from traffic-related tweets. We are interested in four types of fine-grained events; specifically: we are interested to identify \giannis{``when” 
(\ie the exact time that the traffic-related event has happened (as described at the corresponding tweet)),
``where” 
(\ie the location that the traffic-related event has happened,
``what” 
(\ie the type of the incident that has happened, \eg accident, traffic jam)
and the ``consequence” of the aforementioned event (\eg lane blocked)}.
\giannis{We frame the slot filling problem into a sequence labeling task. Given the input sequence of tokens $X = (x_{1}, x_{2}, \cdot \cdot \cdot  , x_{n})$, the goal is to map the input sequence $X$ into a tagged (labeled) sequence $Y = (y_{1}, y_{2}, \cdot \cdot \cdot  , y_{n})$ of the same length, where $n$ is the number of tokens within the sequence. We employ the BIO (beginning-inside-outside) scheme for tagging the sequences and more details about that can be found in~\figref{fig:subtasks}.} \xiangyu{The BIO scheme is a tagging format for labeling tokens according to the positions of the tokens within \giannis{a} chunk. The O tag \giannis{indicates} \giannis{that} the corresponding token is outside of \giannis{the} chunk. The B- prefix before \giannis{the} tag \giannis{indicates} \giannis{that} the corresponding token is \giannis{at} the beginning of \giannis{the} chunk. The I- prefix before \giannis{the} tag \giannis{indicates that} the corresponding token is inside a chunk. \giannis{Thus, there is the constraint} that \giannis{the} tag with the I- prefix should always come after \giannis{the} tag with the B- prefix or the I- prefix \giannis{(of the same type, \eg where).}} 


\subsection{Datasets}
\label{3.2}

\begin{figure}
    \centering
     \resizebox{0.9\columnwidth}{!}{%
    \includegraphics[scale=0.55]{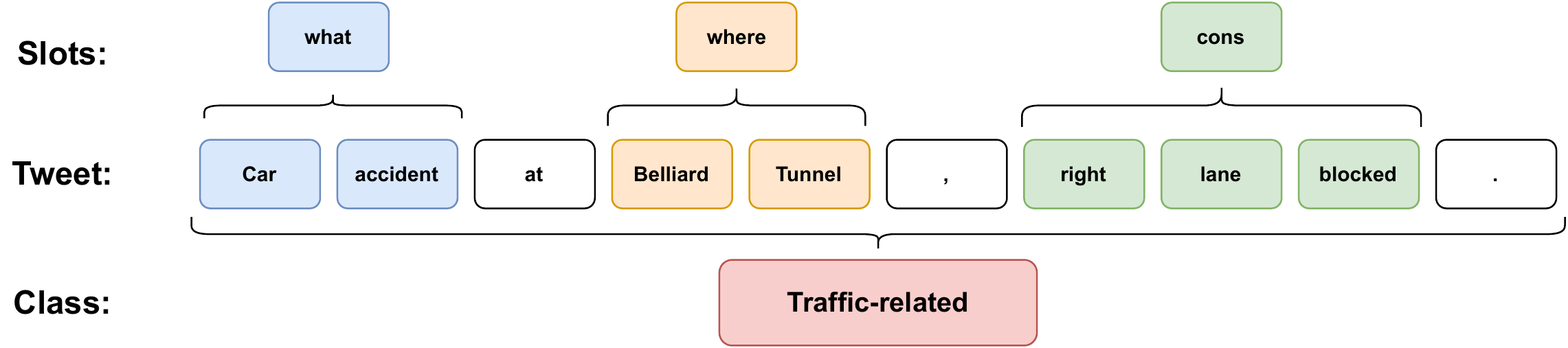}
    \caption{An example of an annotated tweet \giannis{from our dataset}. This is  not  a  fictitious  tweet,  but  rather  a  tweet  that  we  translate  from  Dutch  to English.
    }
    \label{fig:annotated_tweet}}
\end{figure}

\noindent 
\nikos{We have constructed two annotated Dutch datasets from the Twitter stream.} \giannis{The first one is from Belgium, the ``Belgian Traffic Twitter Dataset'' (BE dataset), and the other is from the Brussels capital region, the ``Brussels Traffic Twitter Dataset'' (BRU dataset).}
In fact, the BRU dataset is part of the BE dataset, and we extract the BRU dataset out of the BE dataset to \giannis{study the traffic events in a particular part of the country, the Brussels capital region}.
The tweets within the two datasets range from 2015 to 2019. The two datasets also contain the \giannis{annotations} that consist of two parts: the class of each tweet (whether a tweet is traffic-related or not) and the slots \giannis{of the} fine-grained information (\ie \giannis{`}`when", ``where", ``what", and ``consequence"). 
\giannis{In Figure~\ref{fig:annotated_tweet}, an example of an annotated tweet is illustrated.} 

\begin{table}[t]
\centering
\resizebox{0.5\columnwidth}{!}{%
\begin{tabular}{l|c|c|}
\cline{2-3}
                                          & \multicolumn{1}{l|}{\textbf{BE  dataset}} & \multicolumn{1}{l|}{\textbf{BRU dataset}} \\ \hline
\multicolumn{1}{|l|}{\textbf{Class}}      & \multicolumn{2}{c|}{\textbf{\# of tweets}}                                           \\ \hline
\multicolumn{1}{|l|}{traffic-related}     & 5,386                                    & 3,213                                     \\ \hline
\multicolumn{1}{|l|}{non-traffic-related} & 5,237                                    & 3,313                                     \\ \hline
\multicolumn{1}{|l|}{Total}               & 10,623                                   & 6,526                                     \\ \hline \hline
\multicolumn{1}{|l|}{\textbf{Slot}}       & \multicolumn{2}{c|}{\textbf{\# of slots}}                                            \\ \hline
\multicolumn{1}{|l|}{where}               & 5,305                                    & 3,144                                     \\ \hline
\multicolumn{1}{|l|}{what}                & 5,121                                    & 2,963                                     \\ \hline
\multicolumn{1}{|l|}{when}                & 5,111                                    & 3,030                                     \\ \hline
\multicolumn{1}{|l|}{consequence}         & 3,939                                    & 2,551                                     \\ \hline
\end{tabular}
\caption{Statistics of the BE dataset (\ie the Belgian Traffic Twitter Dataset) and the BRU dataset (\ie the Brussels Traffic Twitter Dataset).}
\label{tab:datasets}}
\end{table}


\noindent \textbf{Data Collection\giannis{:}} Dutch and French are the two most common languages used in Belgium. Since Brussels is an international city, English is also commonly used.
\giannis{Thus,} we decided to track tweets in Dutch, French and English from Belgium. The first step \giannis{in order} to collect a large Twitter dataset from Belgium is to use \giannis{the} Twitter API to harvest tweets from Belgium \giannis{in the specified time period (\ie from 2015 to 2019)}. 
\giannis{Similar to the work of~\citet{dabiri2019developing}, where they establish a list of traffic-related keywords to filter out traffic-related tweets, we established our own traffic-related keywords in Dutch, French, and English.}
Detailed information can be found on \giannis{our} GitHub \giannis{codebase}\footnote{The GitHub repository will be available upon acceptance of the manuscript.}
\giannis{We then use these keywords on the harvested tweets and we obtain potential traffic-related tweets in the aforementioned three languages.}
To investigate the possible distribution of \giannis{these} potential traffic-related tweets regarding the three languages, 
we translate all the non-English tweets into English (using \giannis{the} translators API for Python) and then build a CNN classifier based on the US traffic dataset from~\citet{dabiri2019developing}. 
After manually investigating the results from the CNN classifier, we \giannis{identify} that \giannis{the majority of the} real traffic-related tweets out of \giannis{all} the potential traffic-related tweets \giannis{were coming from Dutch speakers}.
Thus, \giannis{this is why} we focus on building a high-quality Dutch annotated Twitter dataset \giannis{for} Belgium.

\noindent \textbf{Data Annotation\giannis{:}} There are two tasks in the annotation process. The first task is to identify whether a given tweet is traffic-related or not. The traffic-related tweets \giannis{can} report non-recurring events or recurring events. The recurring events are the events that are predictable, such as traffic congestion, traffic delay and daily rush hours. The non-recurring events include unpredictable events such as car accidents, traffic signal problem and disabled vehicles. The second task is to find relevant information \giannis{(\eg ``when", ``where") from tweets identified as traffic-related.}  
In order to easily annotate our dataset, we \giannis{used} the TagTog platform\footnote{\url{https://www.tagtog.net/}}, which is an annotation platform that can improve the annotation experience. Apart from \giannis{using} the annotation platform, we also hired a native Dutch speaker to help us with the annotation. 
\giannis{At the end of the annotation process,} we \giannis{have} also manually checked the annotation results. 
\giannis{The BE dataset contains 10,623 tweets, and the BRU dataset (a part of the BE dataset from the Brussels capital region) contains 6,526 annotated tweets as reported also in~\tabref{tab:datasets}.} \xiangyu{The problem that we are going to address in this paper is not a straightforward problem to solve (e.g., with a predefined set of keywords). This is because when we pre-filter a large fraction of the tweets with a predefined keyword set, these tweets belong to the traffic-related class, however when we annotate them, these tweets belong to the non-traffic related class.}
\section{Proposed Architectures}
\label{method}

\noindent \giannis{In this section, we describe the proposed approaches for solving the two subtasks (\ie text classification and slot filling) either independently or in a joint setting.} 
\giannis{In~\secref{subsec:corecomonents}, we introduce the three main components (\ie CNN, LSTM, and BERT) that are mainly exploited by the independent (\ie the two subtasks are considered separately, see~\secref{subsec:independent_models}) and the joint (\ie the two subtasks are considered in a joint setting, see~\secref{subsec:joint_models}) models for solving the traffic event detection problem.}

\subsection{Core Components}
\label{subsec:corecomonents}

\subsubsection{Neural-based Methods}
\noindent 
\giannis{\textbf{Convolutional \giannis{N}eural \giannis{N}etworks (CNNs)}: \giannis{CNNs have been mainly exploited in computer vision tasks (see \eg image classification~\citep{Krizhevsky2012ImageNetCW,Girshick2015FastR,He2020MaskR}, semantic segmentation~\citep{10.1145/3272127.3275050}, image super-resolution~\citep{Zhang2020CascadeCN}, etc)}. However, CNNs have also been extensively used in NLP for tasks, such as text classification~\citep{Kim2014ConvolutionalNN}, sequence labeling~\citep{xu2018double}, etc. due to their ability to extract \giannis{$n$-gram features}. This kind of models consists of a number of  convolutional filters (of various sizes) that are applied on top of the embedding layer. 
In this work, CNNs are mainly used as \giannis{a} core component for the text classification \giannis{sub}task.} 

\noindent \giannis{\textbf{Long Short-Term Memory (LSTMs)}: LSTMs, a variant of Recurrent Neural Networks (RNNs)~\citep{Hochreiter1997LongSM}, can handle data of sequential nature (\ie text) and showcase state-of-the-art performance in a number of NLP tasks (see~\eg text classification~\citep{Zhou2015ACN}, sequence labeling~\citep{lu2019sc}, fact checking~\citep{rashkin2017truth,bekoulis2020review}).} \nikos{RNNs suffer from the vanishing gradient problem which harms convergence when dealing with long input sequences. LSTMs use modifications, such as the cell state 
\giannisrtwo{(\ie the memory of LSTM)} to overcome the vanishing gradient problem. 
The standard LSTM processes the input sequence from left to right, and for each input token, LSTMs produce a hidden state as output that takes the previous input tokens into account.
LSTMs can also be applied from right to left and thus bidirectional LSTMs (BiLSTMs) can obtain bidirectional information for each input token.}

\subsubsection{BERT}
\label{subsubsec:bert}

\noindent \giannis{Bidirectional Encoder Representations from Transformers (BERT)~\citep{bert} is a Transformer-based language representation model~\citep{vaswani2017attention}, where multiple Transformer encoders are stacked the one on top of the other, and are pre-trained on large corpora.
For each input sequence, a special classification token~\clstag~is added at the beginning and a special token~\septag~is added at the end of each sentence. The outputs of the BERT model are high-level deep bidirectional contextual representations of the input tokens. }\\

\noindent\textbf{BERT-based Pre-trained Models for \giannis{the} Dutch Language}

\noindent Since both \nikos{constructed} datasets are in Dutch, we \giannis{use four BERT-based models that are pre-trained on Dutch data.} The four pre-trained Dutch models \giannis{are the following:} BERTje~\citep{de2019bertje}, RobBERT~\citep{delobelle2020robbert}, mBERT~\citep{bert}, and XLM-RoBERTa~\citep{conneau2019unsupervised}. 

\noindent \textbf{BERTje} is a Dutch BERT model that is pre-trained on a large and diverse Dutch dataset of 2.4 billion tokens \xiangyu{from Dutch Books, TwNC~\citep{Ordelman2007TwNCAM}, SoNaR-500~\citep{Oostdijk2013TheCO}, Web news and Wikipedia}.

\noindent \textbf{RobBERT} is a Dutch language model based on the \giannis{Robustly Optimized BERT method (RoBERTa~\citep{liu2019roberta})} and is pre-trained on the Dutch \giannis{part} of the OSCAR corpus with 6.6 billion words.

\noindent \textbf{mBERT} is a multilingual BERT model, which \giannis{also includes the Dutch language.} The Dutch part of the model is \giannis{only pre-trained} on Dutch Wikipedia text.

\noindent \textbf{XLM-RoBERTa} is a multilingual model trained on 100 different languages\giannis{,} which include\giannis{s} Dutch and is based on RoBERTa.


\subsection{Independent Models}
\label{subsec:independent_models}

\noindent In this \giannis{subsection}, we \giannis{describe} the models that address the two subtasks independently.

\subsubsection{Text Classification}
\label{subsubsec:textclassification}

\noindent \textbf{CNN}: 
\giannis{The} CNN model \giannis{consists of} four parts: \giannis{the word embeddings layer}, a convolutional layer, a max pooling layer and a fully connected layer. 
\giannis{Specifically, at the word embeddings layer, the input tokens are mapped to word embeddings (\ie word vectors).} Then the convolutional layer extracts \giannis{$n$}-gram features from the input and those features are further processed by \giannis{a} max pooling layer. Finally, the processed features are passed to the fully connected layer to make \giannis{class} predictions \giannis{for} the input sequence.

\noindent  \textbf{LSTM}: 
This LSTM model consists of \giannis{three} parts\giannis{:} \giannis{the} word embeddings \giannis{layer}, a BiLSTM layer, and a softmax layer. The input sequence is converted into pre-trained word embeddings. Then, the word embeddings are processed by the BiLSTM layer\giannis{, and t}he final hidden states of the forward and the backward LSTMs are concatenated. The\giannis{n the concatenated hidden states are} passed to a softmax layer to predict the class of the input sequence.

\begin{figure}[t!]
    \centering
    \includegraphics[scale=0.7]{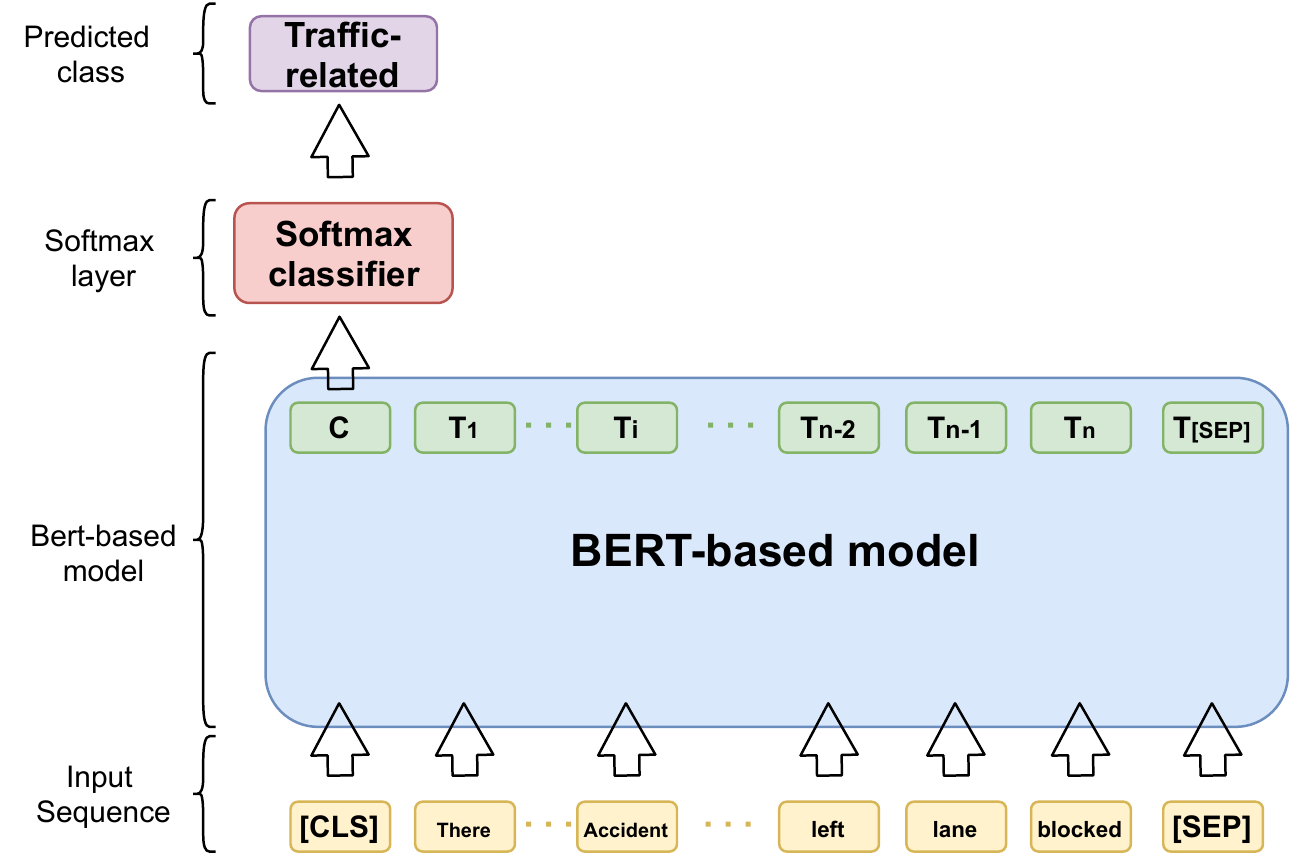}
    \caption{BERT-based model for text classification. The model \giannis{consists} of three parts, the input sequence \giannis{(}yellow rectangles\giannis{)}, the BERT-based model \giannis{(}blue rectangle\giannis{)}, and the softmax layer \giannis{(}red rectangle\giannis{)}. The green rectangles represent the hidden states of the corresponding input tokens. The class of the input sequence is predicted by applying a softmax layer on top of the hidden state of the \clstag~token. The predicted class is shown in the purple rectangle. }
    \label{fig:bert-cls}
\end{figure}
\noindent  \textbf{BERT-based Model}:
 \giannis{T}he hidden state of the~\clstag~tag encodes the information of the whole input sequence~\citep{bert}. 
 \giannis{Thus}, \giannis{we use} the~\clstag~representation to \giannis{represent the whole sentence for the} text classification \giannis{task}. 
 The architecture of the BERT-based model is shown in \Figref{fig:bert-cls}. To predict the class of the input sequence, a softmax layer is applied on top of the hidden state of the \clstag~token. The equation is:
\begin{equation}
\label{eqn:txtcls}
    y^c = \text{softmax} (W^ch_{[cls]}+b^c)
\end{equation}
\noindent where $y^c$ is the prediction of the input sequence, $W^c$ is the weight matrix, $h_{[cls]}$ is the hidden state of the special \clstag~token, and $b^c$ is the bias vector.

\subsubsection{Slot Filling}
\label{subsubsec:slotfilling}

\noindent \yang{We treat the slot filling task as a sequence labeling problem. For each token within a tweet, we assign the slot label with the highest probability.}

\noindent \textbf{LSTM}: 
\giannisrtwo{This model has a similar structure to the LSTM model presented for the text classification task in~\secref{subsubsec:textclassification}. However, in the slot filling case, the concatenated BiLSTM hidden states for each input token are used to predict the tag for token.}

\noindent  \textbf{LSTM + CRF}: 
\giannis{Instead of using a softmax layer on top of the LSTM model that independently predicts the tag for each input token}, a conditional random field (CRF) layer~\citep{lample-etal-2016-neural} is employed to \giannis{capture the relationships between neighboring tokens}.

\begin{figure}[t!]
    \centering
    \includegraphics[scale=0.7]{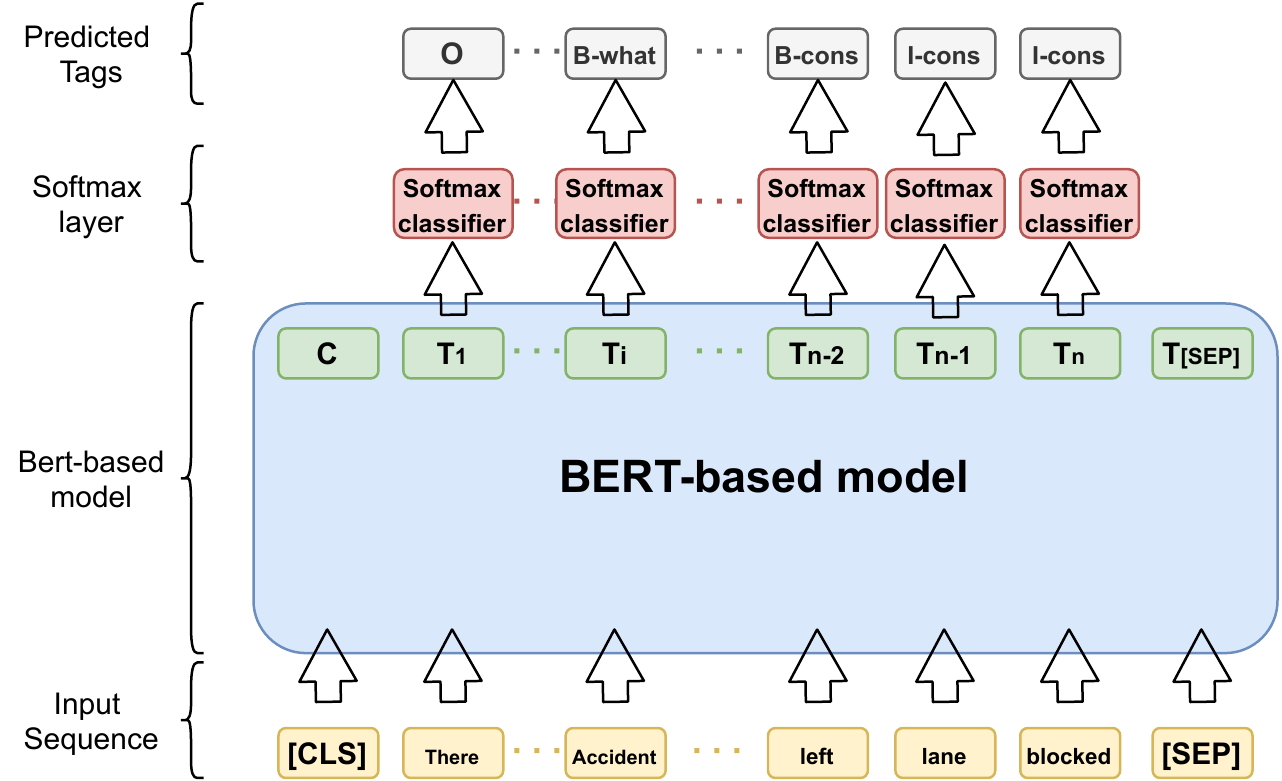}
    \caption{BERT-based model for slot filling. The model consists of the input sequence \giannis{(}yellow rectangles\giannis{)}, the BERT-based model \giannis{(}blue rectangle\giannis{)}, and a softmax layer \giannis{(}red rectangles\giannis{)}. Each green rectangle contains the hidden state of the corresponding input token. Each grey rectangle is the predicted tag for the corresponding token.}
    \label{fig:bert-ner}
\end{figure}

\noindent  \textbf{BERT-based Model}:
 \Figref{fig:bert-ner} shows the structure of the BERT-based model for slot filling. The hidden states of the input tokens are used for labeling the tokens. A softmax classifier is added over the hidden state of each corresponding input token to predict the corresponding tag. 
 \giannis{The BERT-based models use the WordPiece tokenizer~\citep{Wu2016GooglesNM} and can partition one word into several sub-tokens according to the vocabulary of the tokenizer.
 However, the model itself should output one prediction per word/token, and each token might be split into several sub-tokens. To handle this issue, we only make predictions for the first sub-token of each token, in the case that a token has been split into multiple sub-tokens (by the WordPiece tokenizer), or for the token itself, in the case that the whole token has been retained.
 }
 The equation for \giannis{the slot filling task} is:
\begin{equation}
    y^s_i = \text{softmax}(W^sh_i + b^{s}),~i\in 1...N
\end{equation}
\noindent where $y^s_i$ is the tag prediction of the token $x_{i}$, $W^s$ is the weight matrix, $h_i$ is the hidden state of the first sub-token of $x_{i}$, and $b^{s}$ is the bias vector. 

\begin{figure}[t!]
    \centering
     \resizebox{0.8\columnwidth}{!}{
    \includegraphics[scale=0.75]{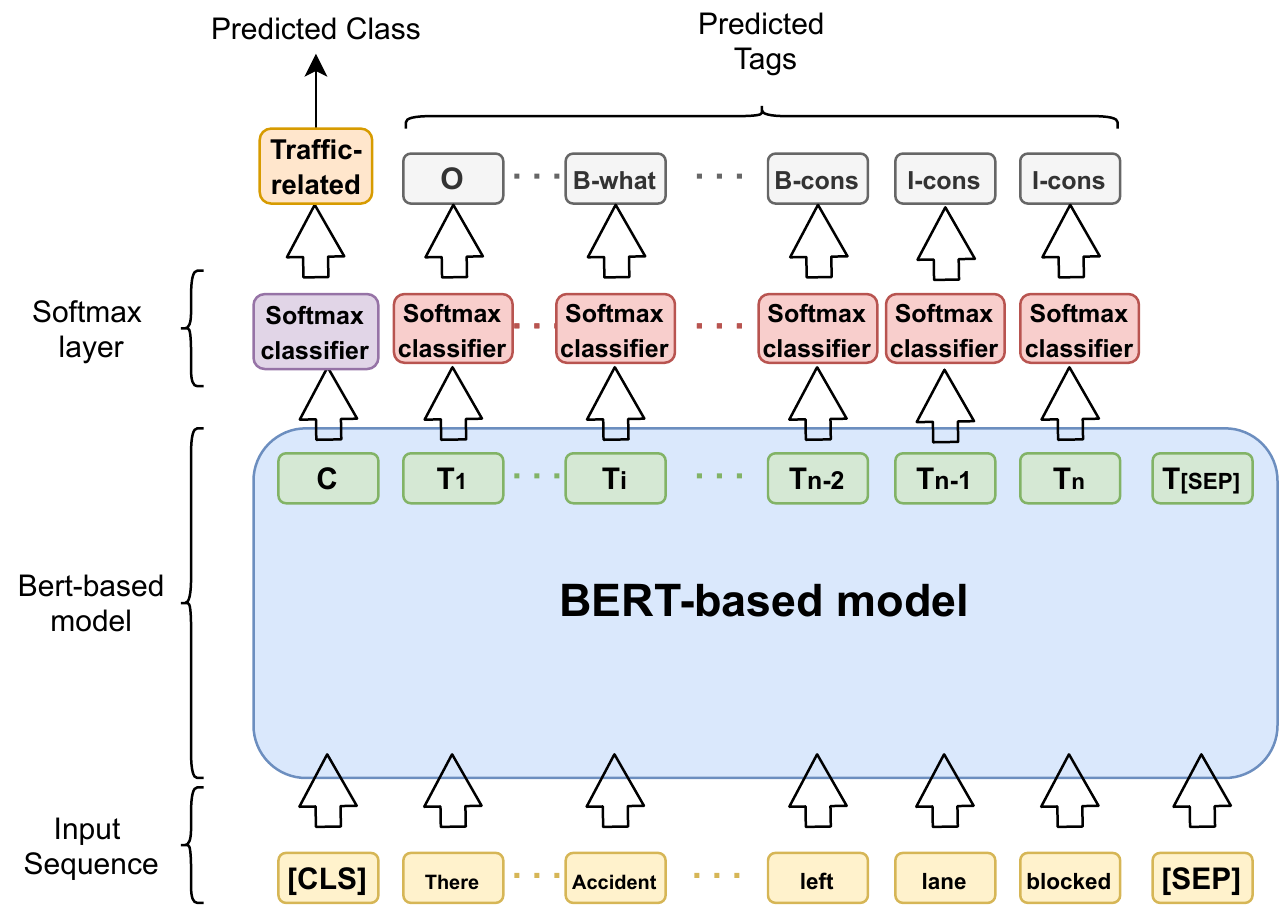}
    \caption{Joint BERT. 
    \giannis{The model consists of the input sequence (yellow rectangles), the BERT-based model (blue rectangle), and softmax layers (purple or red rectangles).}
    The orange rectangle is the predicted class for the input sequence, and each grey rectangle is the predicted tag for the corresponding input token. For this model, the text classification and slot filling tasks are jointly trained.}
    \label{fig:joint_bert_arch}}
\end{figure}

\subsection{Joint Models}
\label{subsec:joint_models}

\noindent \giannis{In this subsection, w}e describe the joint models for the two subtasks\giannis{:}

\noindent \textbf{Slot-Gated}~\citep{goo2018slot}: 
Built on top of a\giannis{n} attention-based BiLSTM \giannis{architecture}, the Slot-Gated model has a slot-gated mechanism, which is designed to learn the relationship between the intent and \giannis{the} slot context vectors to improve the performance of \giannis{the} slot filling \giannis{task}.

\noindent \textbf{SF-ID Network}~\citep{haihong2019novel}: 
Based \giannis{also on BiLSTMs}, the SF-ID network can directly establish connections \giannis{between} the intent detection and \giannis{the} slot filling \giannis{subtasks}. \giannis{In addition}, an iteration mechanism is designed to enhance the interrelated connections \xiangyu{between the intent and the slots}.  
\begin{figure}[t!]
    \centering
    \resizebox{0.75\columnwidth}{!}{
    \includegraphics[scale=0.85]{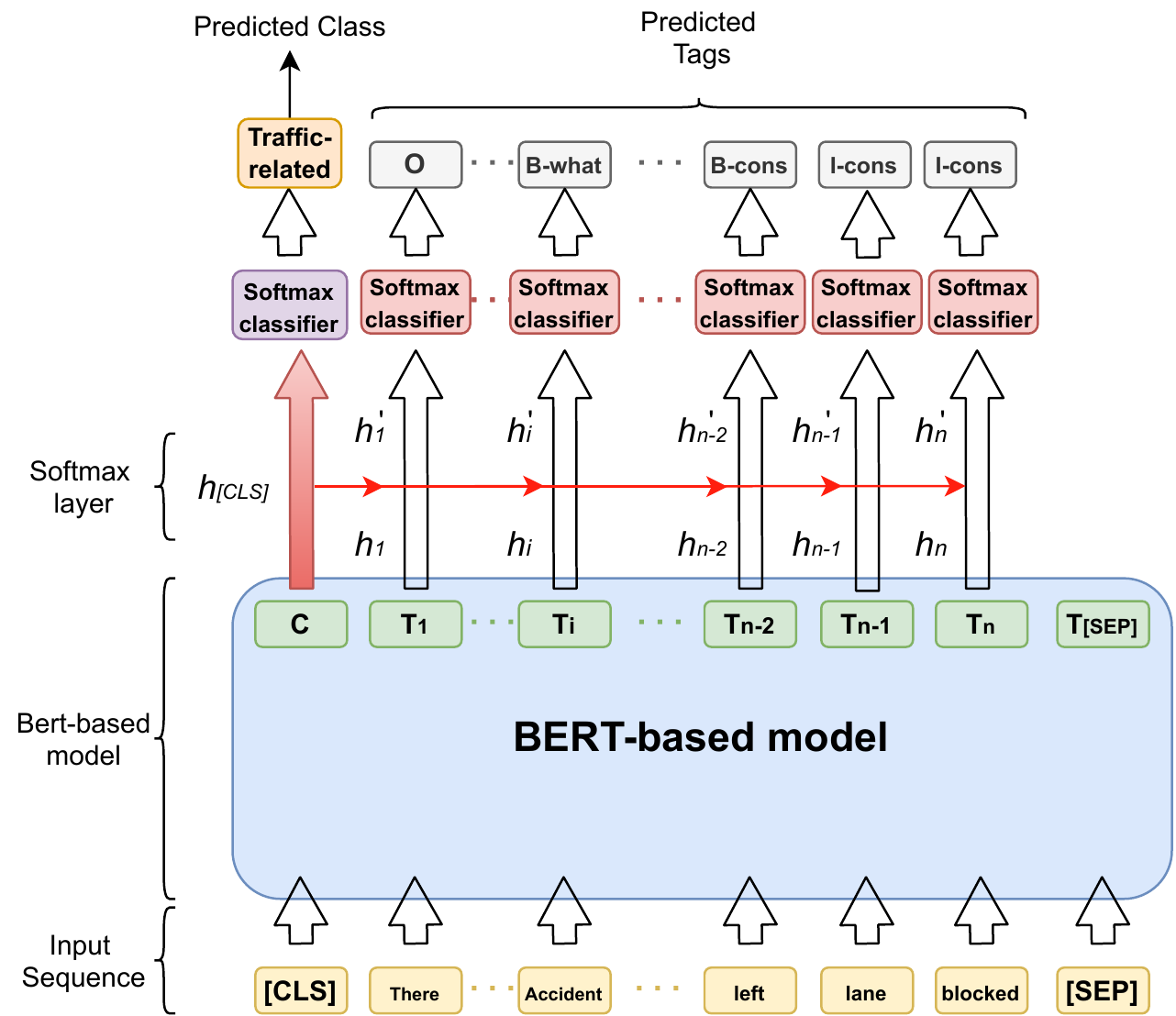}
    \caption{
    Joint Enhanced BERT-based model for joint text classification and slot filling. \giannis{This model is similar to the joint BERT architecture but the \clstag~ hidden state is concatenated to each of the hidden states of the tokens of the sentence. The model consists of the input sequence (yellow rectangles), the BERT-based model (blue rectangle), and softmax layers (purple or red rectangles).} The red arrows connect the sentence contextual information with each non-special token. The orange rectangle is the predicted class of the input sequence, and each grey rectangle is the predicted tag for the corresponding input token. \giannis{Note that} the text classification and slot filling tasks are jointly trained. }
    \label{fig:bert-joint}
    }
\end{figure}

\noindent \textbf{Capsule-NLU}~\citep{zhang2019joint}: 
This model uses a dynamic routing-by-agreement schema to tackle \giannis{the} intent detection and \giannis{the} slot filling \giannis{subtasks}. \xiangyu{As a result,} the model is able to preserve the hierarchical relationship among \giannis{the two subtasks}.

\noindent \textbf{Joint BERT}~\citep{chen}: 
\giannis{The} model\nikos{, depicted} in~\Figref{fig:joint_bert_arch}, \giannis{uses the hidden state of the~\clstag~token to perform the intent detection \giannis{subtask},} and \giannis{the} slot filling \giannis{subtask} on the hidden states of \giannis{the} other non-special tokens \giannis{(\ie the tokens of the sentence)}. The two tasks are jointly trained by adopting a joint loss function.

\noindent \textbf{Joint Enhanced BERT-based Model}: 
 \yang{While the previous models have already been proposed in the literature, in what follows we describe our proposed model, the Joint Enhanced BERT-based model, for the joint task.}
~\Figref{fig:bert-joint} shows the structure of our proposed model. 
 \giannis{This models takes into account the two following facts:}
 (i) \giannis{t}he class of a tweet (\giannis{\ie} traffic-related or not) is relevant \giannis{to the slot filling problem} (\giannis{\ie identify more} fine-grained information\giannis{, \eg} ``when", ``where"), and vice versa\giannis{, and (ii) t}he hidden state of the~\clstag~\giannis{contains} the information of the entire input sequence. Compared to the joint BERT model~\citep{chen} which only trains the two tasks together using a joint loss without modeling the relationships between them, we incorporate the information of the whole input sequence into each token for \giannis{improving the performance of the model}. For each non-special token, we concatenate its hidden state with the hidden state of the~\clstag~\giannis{tag}. The concatenation is used to predict the label for the token \giannis{similar to the way that we were using the hidden state of the token in the standard Joint BERT model (described above):}

\begin{equation}
    h_{i}^{'} = [h_{i}, h_{[cls]}]
\end{equation}
\begin{equation}
    y^{s'}_i = \text{softmax}(W^{s'}h_i^{'} + b^{s'}), i\in 1...N
\end{equation}

\noindent where $h_{i}^{'}$ is the concatenation of the hidden states of the $i_{th}$ token and the~\clstag~token, $y^{s'}_i$ is the tag prediction of the token $x_{i}$, $W^{s'}$ is the weight matrix, $h_i^{'}$ is the hidden state of the first sub-token of $x_{i}$, and $b^{s'}$ is the bias vector.
The goal of this joint enhanced model is to maximize the conditional probability:
\begin{equation}
    p(y^c, y^{s'}|X) = p(y^c|X)\prod_{i=1}^Np(y^{s'}_i|X)
\end{equation}
\noindent where $y^c$ is the prediction of the input sequence,  $y^{s'}_i$ is the tag prediction of the token $x_{i}$, and $X$ is the input sequence. We use Equation~\ref{eqn:txtcls} for predicting the class of a tweet. 

\section{Experiments and Results}
\label{experiments}

\noindent \giannis{In this section, we describe, (i) the evaluation metrics for all the experimented methods, and (ii) the experimental settings for the various models. Finally, we evaluate the performance of the proposed architectures for solving the traffic event detection problem and we discuss the results.}

\subsection{Evaluation Metrics}
\label{subsec:eval_metric}
\noindent We adopt three evaluation metrics for evaluating \giannis{the} different models on \giannis{the constructed datasets}. 
\giannis{We use the F$_1$ score for the two subtasks of text classification and slot filling, denoted as F$_{1c}$ and F$_{1s}$, respectively.}
For the joint text classification and slot filling, the sentence-level semantic frame accuracy (SenAcc) \giannis{score is calculated, and indicates the proportion of sentences (out of all sentences) that have been correctly classified. \yang{I}n order for a sentence to be correct both the class and the slots of the sentence should be identified correctly.}

\subsection{Experimental Settings}
\label{subsec:exp_setting}

\noindent 
\giannis{We randomly split the two datasets (\ie BRU and BE); specifically, we keep 60\% for training, 20\% for validation, and 20\% for the test sets for each of the datasets. }
\giannis{Since the} URLs do not provide useful information for traffic events, we remove all the URLs from the tweets \giannis{similar to the work of}~\cite{dabiri2019developing}. For non-BERT-based models, we use the 160-dimensional Dutch word embeddings 
\giannis{called} Roularta~\citep{tulkens2016evaluating}. For \giannis{the} BERT-based models, the batch size is set to 32 or 64. The dropout rate is 0.1. The number of epochs is selected from \giannis{the values} 10, 15, 20, 25, 30, 40. Adam~\citep{Kingma2015AdamAM} is used to optimize the model parameters with an initial learning rate of 1e-4\giannis{, and} 5e-5 for the joint and \giannis{the independent} models, respectively. 
\xiangyu{For the CNN model, the filters in the convolutional layer have a size of $n \times 160$, where $n = [3,4,5]$. We apply dropout with a rate of 0.5 on the outputs \giannis{of} the max pooling layer. 
\giannis{In} the LSTM model for text classification, the hidden dimension of the BiLSTM layer is set to 256. Dropout is applied on the output of the BiLSTM layer with a rate of 0.5. 
For these two models (\ie the CNN model and the LSTM model for text classification), the Adam optimizer~\citep{Kingma2015AdamAM} with a learning rate of 0.001 is used to update the model parameters.
For the LSTM and the LSTM + CRF models for slot filling, the hidden dimension of the BiLSTM layer is 100. The stochastic gradient decent algorithm with a learning rate of 0.015 is used for updating the model parameters. The dropout rate is set to 0.5 for the output of the BiLSTM layer.
For the Slot-Gated, SF-ID Network, and Capsule-NLU models, we use their default settings as described in their papers~\citep{goo2018slot, haihong2019novel, zhang2019joint}.}

\subsection{Results}
\label{subsec:results}
\subsubsection{Independent Models}
\label{sec:res:independent_models}

\noindent\giannis{Table~\ref{tab:indenpendent} \nikos{reports} the performance of \giannis{the} different models for the tasks of text classification and slot filling on the BRU and the BE datasets, \giannis{where the two tasks are considered independently}. \giannis{We also report transfer learning experiments (see the BRU$\rightarrow$BE column in Table~\ref{tab:indenpendent} and the transfer learning part of this Section for more details).} For the text classification task,  
based on the results, we observe that the BERTje model performs best in terms of F$_{1c}$ score on both datasets. In particular, BERTje scores 98.56\% and 97.31\% on the BRU and the BE datasets, respectively.}
\begin{table}[!t]
\centering
\resizebox{0.9\columnwidth}{!}{%
\caption{The results of the different models for the text classification and the slot filling tasks on the two datasets as well as the generalization ability of the best performing BRU models for text classification and slot filling on \giannis{the} BE dataset (except the BRU part\giannis{, see the BRU$\rightarrow$BE column}) in terms of \Fone~score. \Fonec~and \Fones~indicate the \Fone~score for the text classification and the slot filling tasks, respectively. \giannis{\yang{T}he two tasks are considered independently.}}
\label{tab:indenpendent}
\begin{tabular}{@{\extracolsep{4pt}}cccccccc@{}}
\toprule
                                      &             & \multicolumn{3}{c}{\Fonec}  & \multicolumn{3}{c}{{\Fones}}\\ 
\cline{3-5}  
\cline{6-8}
                                      & Model             & BRU & BE & BRU$\rightarrow$BE  & BRU & BE & BRU$\rightarrow$BE\\ 
\midrule
\midrule
\parbox[c]{5mm}{\multirow{3}{*}{\rotatebox[origin=c]{90}{\parbox{1.4cm}{\centering Neural-based}}}}
\multirow{3}{*}{}                     & CNN         & 97.96  & 96.73 & 95.97    & - & - & -\\
                                      & LSTM        & 97.95  & 96.53 & \textbf{96.56}    & 97.97     & 96.44  & 93.03\\ 
                                      & LSTM + CRF       & -  & -  & -   & \textbf{98.18} & \textbf{97.22} & \textbf{95.24}\\
\midrule
\parbox[c]{5mm}{\multirow{4}{*}{\rotatebox[origin=c]{90}{\parbox{1.4cm}{\centering BERT-based}}}} & BERTje      & \textbf{98.56}  & \textbf{97.31}   & 94.53    & 97.79         & 97.18 & 92.24\\
                                      & RobBERT     & 98.14  & 96.66  & 95.81   & 96.40          & 96.10  & 77.46\\
                                      & mBERT       & 98.35  & 97.08  & 94.82   & 97.98          & 96.87  & 90.03\\
                                      & XLM-RoBERTa & 98.20  & 96.22  & 95.44   & 96.76          & 96.52  & 80.31\\
\bottomrule
\end{tabular}}
\end{table}
\nikos{The BERT-based} models perform better or on par with the non-BERT-based models on both datasets. The reason could be that BERT-based models can better encode the contextual information of the whole input sequence. On the BRU dataset, all the models \giannis{achieve an accuracy score of around} 98\%, \giannis{and for the BE dataset the score is around 96\%}.
On the slot filling task, the LSTM + CRF \giannis{model} outperforms all the other models on both datasets, \giannis{and the score is} 98.18\% on the BRU dataset and 97.22\% on the BE dataset. On the BRU dataset, \giannis{the rest of the} models achieve an~\Fone~score of around 97\%. On the BE dataset, the BERTje has a similar~\Fone~score (97.18\%) to the LSTM + CRF \giannis{model}. The rest \giannis{of the} models \giannis{achieve an~\Fone~score of around 96\%}. On both datasets, the BERT-based models can perform on par with the non-BERT-based models. 
\giannis{We} notice that the LSTM + CRF models have better performance compared to the models \giannis{that rely only on LSTMs on both datasets}. This is due to the fact that the CRF layer can \giannis{take into account the relationships of the outputs between sequential tokens (\ie a B-where tag cannot be followed by an I-when tag).}

\noindent \textbf{Transfer Learning:} 
Table~\ref{tab:indenpendent} shows the \giannis{results} of the best performing BRU models \giannis{on the BE dataset (except \giannis{for} the BRU part \giannis{of the dataset, see the BRU$\rightarrow$BE column}) for} the tasks of text classification and slot filling in terms of \Fone~score. 
For \giannis{the} text classification \giannis{task}, the results indicate that \giannis{the} non-BERT based models generalize better on this task with 96.56\% for the LSTM \giannis{model} and 95.97\% for the CNN \giannis{model} in terms of classification \Fonec~score. However, the BERT-based models still generalize well with around 95\% classification \Fonec~score. 
For \giannis{the} slot filling \giannis{task}, \nikos{the} LSTM + CRF has the best generalization ability. \xiangyu{The reason could be that the CRF layer can learn the relationships between the neighboring tokens.} The RobBERT \giannis{model} \nikos{achieves} the worst generalization performance. \nikos{This is mainly because the RoBERTa model needs more data for being fine-tuned.}



\begin{table}
\centering
\resizebox{0.95\columnwidth}{!}{%
\caption{The results of different models for joint text classification and slot filling on the two datasets as well as the generalization ability of the best performing trained BRU joint models \giannis{on} the BE dataset (except \giannis{for} the BRU part\giannis{, see the BRU$\rightarrow$BE column}) in terms of \Fone~score and sentence-level semantic frame accuracy (SenAcc). \Fonec~and \Fones~indicate the \Fone~score for the text classification and the slot filling tasks, respectively.}
\label{tab:joint}
\begin{tabular}{@{\extracolsep{4pt}}cccccccccccc@{}}
\toprule
                                      & \multirow{2}{*}{Model} & \multicolumn{3}{c}{BRU}                          & \multicolumn{3}{c}{BE}                      & \multicolumn{3}{c}{BRU$\rightarrow$BE}      \\ 
\cline{3-5} 
\cline{6-8}
\cline{9-11}
                                      &                        & \Fonec        & \Fones         & SenAcc         & \Fonec        & \Fones         & SenAcc          & \Fonec        & \Fones         & SenAcc\\ 
\midrule
\midrule
\parbox[c]{5mm}{\multirow{3}{*}{\rotatebox[origin=c]{90}{\parbox{1.4cm}{\centering LSTM-based}}}}
\multirow{3}{*}{}                     & Slot-Gated             & 95.09          & 97.10          & 92.76          & 96.23          & 96.13          & 92.85          & 94.17                   & 76.20                                       & 66.29 \\
                                      & SF-ID Network          & 97.23          & 97.03          & 94.10          & 96.07          & 96.67          & 93.84           & 84.78                   & 83.29                                       & 69.03\\
                                      & Capsule-NLU            & 97.75          & 97.72          & 95.79          & 96.71          & 96.76          & 94.21           & 95.92                   & 91.45                                       & 87.60\\ 
\midrule
\parbox[c]{5mm}{\multirow{4}{*}{\rotatebox[origin=c]{90}{\parbox{1.4cm}{\centering Joint BERT}}}} 
                                      & BERTje                 & 98.34          & 97.31          & 95.79          & 96.71          & 96.81          & 93.93            & 94.18                   & 92.11                                       & 86.23\\
                                      & RobBERT                & 97.90          & 96.46          & 94.10          & 96.31          & 96.21          & 92.90           & 94.62                   & 74.15                                       & 70.88\\
                                      & mBERT                  & 97.89          & 97.42          & 95.71          & 96.95          & 96.63          & 94.26           & 94.58                   & 92.58                                       & 87.01\\
                                      & XLM-RoBERTa            & 97.97          & 96.53          & 94.10          & 96.51          & 96.24          & 93.13           & 92.99                   & 76.09                                       & 53.14\\ 
\midrule
\parbox[c]{5mm}{\multirow{4}{*}{\rotatebox[origin=c]{90}{\parbox{1.4cm}{\centering Enhanced BERT}}}}
                                      & BERTje                 & \textbf{98.49} & 97.52          & 95.87          & \textbf{97.13} & \textbf{97.10} & \textbf{94.63}  & 95.27         & 93.78                             & 88.67\\
                                      & RobBERT                & 98.12          & 96.65          & 94.72          & 96.53          & 96.40          & 93.41           & 94.96                   & 80.00                                       & 74.84\\
                                      & mBERT                  & 97.92          & \textbf{98.29} & \textbf{96.40} & 96.82          & 96.78          & 94.59           & \textbf{96.56}                   & \textbf{94.13}                                       & \textbf{89.94}\\
                                      & XLM-RoBERTa            & 98.20          & 96.65          & 94.56          & 96.26          & 96.18          & 92.76           & 95.76                   & 74.21                                       & 72.57\\
\bottomrule
\end{tabular}
    }
\end{table}
\subsubsection{Joint Models}
\noindent Table~\ref{tab:joint} shows the \giannis{performance} of the joint models (Slot-Gated, SF-ID Network, Capsule-NLU, Joint BERT-based, Enhanced Joint BERT-based) on the BRU and \giannis{the BE datasets in terms of \Fone~score 
and semantic accuracy (SenAcc)}. \giannis{\yang{I}n the joint setting,} the higher the semantic frame accuracy score, the better the joint model performance. On the BRU dataset, the Enhanced Joint mBERT achieves the best semantic accuracy score \giannis{(\ie 96.40\%)} as well as the best \Fones~score \giannis{(\ie 98.29\%)}. The Enhanced Joint BERTje \giannis{model} has the best \giannis{\Fonec~score (\ie 98.49\%)}. 
\giannis{However, on the BE dataset, the Enhanced Joint BERTje is the best performing model and it achieves scores of 97.13\% in terms of \Fonec~score, 97.10\% in terms of \Fones~score, and 94.63\% in terms of SenAcc.}
Among the three non-BERT-based models, the Capsule-NLU model \giannis{achieves} the best results on both datasets and it performs on par with \giannis{the} BERT-based models. 
\giannis{That is due to ability of the model to capture hierarchical semantic relationships between the two subtasks. } 
As we \giannis{observe in} the results, the Enhanced Joint BERT-based models have better performance than the Joint BERT-based models. \giannis{This indicates that the proposed Enhanced models are more effective for the joint task, mainly because the model can incorporate the entire sentence information into each token.
Thus, we conclude that the Enhanced Joint BERT-based models, especially the BERTje and mBERT variants, can be used as strong baselines for the joint text classification and slot filling tasks for the traffic event detection problem in Belgium and in the Brussels capital region.} 

\noindent \giannis{\textbf{Transfer Learning:}} Table~\ref{tab:joint} shows the generalization ability of the best performing BRU models for \giannis{the joint task of text classification and slot filling}. \nikos{The Enhanced Joint BERT-based models} \giannis{obtain the best generalization performance not only compared to the Joint BERT-based models, but also to the other models proposed for solving the task. This implies that the Enhanced Joint BERT-based model improves the model generalization ability by taking the whole information about the sentence for each token into account. Due to its architecture, the Capsule-NLU model can preserve the hierarchical relationships between the two subtasks. That way, it is able to outperform all the other LSTM-based models and perform on par with the Joint BERT-based models.} 

\subsection{Performance of the Independent and the Joint Models}
\noindent \nikos{The results in Tables~\ref{tab:indenpendent} and~\ref{tab:joint} indicate that} \giannis{the joint models perform on par with the independent models on the two subtasks. The benefit of the joint training is not evident in that particular problem as it was in the case of other NLP problems (\eg \cite{miwa2016end,hashimoto-etal-2017-joint}). We hypothesize that this is because the performance of all the models (joint and independent) for the two subtasks is already high in terms of \Fone~score and SenAcc. Thus, there is very little room for improvement for the joint models for the traffic event detection problem. However, one advantage of training a joint model (for that particular problem) is that these models can be deployed easier since we have to train only one model instead of two models, \ie one for each subtask.}

\giannis{The performance of the various models is high, and this could be potentially explained by the large number of tweets related to traffic that are coming from various organizations. Although, the tweets (from the organizations) are not structured, they follow specific patterns (\eg the ``where'' slots are usually followed by the ``when'' slots). This is also the case for the tweets presented in the work of~\citet{dabiri2019developing}, where the authors have focused their study in the US, and the text classification performance in their work was also high. It is worthwhile mentioning though that this is not a disadvantage since it is crucial to have high \Fone~scores in a problem, where the data could be used in real time applications (\eg send alert notifications to users).}

\section{Conclusion and Future Work}
\label{conclusion}
\noindent In this work, we define the \giannis{new} problem of detecting traffic-related events on Twitter \giannis{that consists of} two subtasks: (i) identify whether a tweet contains traffic-related events as a text classification task, and (ii) identify fine-grained traffic-related information \giannis{(\eg ``when’’, ``where”)} as a slot filling task. \giannis{We also} publish \giannis{the two constructed Dutch} traffic Twitter datasets to promote \giannis{further} research \giannis{on} detecting traffic-related events from social media. 

\giannis{We extensively experiment with several architectures for resolving the two tasks either separately or in a joint setting. In this paper, we investigate several baseline methods for solving the two subtasks, and we propose a certain modification in the BERT-based model (see our Enhanced Joint BERT-based model in~\secref{subsec:joint_models}), where we take into account the entire sentence information for each token to further improve the performance of the model. Experimental results indicate that the proposed model is able to outperform all the other studied architectures for solving the task at hand \giannisrtwo{jointly} in several scenarios, \ie datasets from Belgium (BE), and the Brussels capital region (BRU), and the transfer learning experiment BRU$\rightarrow$BE.} 

The current traffic event detection models are mainly built for Dutch tweets. As there are three official languages in Belgium, \giannis{as future work, we aim to work} on multilingual models which can handle \giannis{various} languages. 
\giannis{That way,} we expect that \giannis{these} models can also be \giannis{exploited} to detect traffic-related events in other countries. 
\giannis{Moreover, since Twitter is} \nikos{a noisy source of information}, \giannis{building a language model that is based on a large Twitter corpus, similar to the work of~\cite{muller2020covid}, can potentially help us to improve the performance of our models.}

\section*{Acknowledgements}

\noindent This work has been supported by the MobiWave Innoviris Project.

\bibliography{sample}

\end{document}